\documentclass{article}

\usepackage{arxiv}

\usepackage[utf8]{inputenc} 
\usepackage[T1]{fontenc}    
\usepackage{hyperref}       
\usepackage{url}            
\usepackage{booktabs}       
\usepackage{amsfonts}       
\usepackage{nicefrac}       
\usepackage{microtype}      
\usepackage{lipsum}

\makeatletter
\@ifundefined{KOMAClassName}{
  \IfFileExists{parskip.sty}{%
    \usepackage{parskip}
  }{
    \setlength{\parindent}{0pt}
    \setlength{\parskip}{6pt plus 2pt minus 1pt}}
}{
  \KOMAoptions{parskip=half}}
\makeatother
\NewDocumentCommand\citeproctext{}{}

\makeatletter
 \let\@cite@ofmt\@firstofone
 \def\@biblabel#1{}
 \def\@cite#1#2{{#1\if@tempswa , #2\fi}}
\makeatother
\newlength{\cslhangindent}
\newlength{\csllabelwidth}
\newenvironment{CSLReferences}[2] 
 {\begin{list}{}{%
  \setlength{\itemindent}{0pt}
  \setlength{\leftmargin}{0pt}
  \setlength{\parsep}{0pt}
  \ifodd #1
   \setlength{\leftmargin}{\cslhangindent}
   \setlength{\itemindent}{-1\cslhangindent}
  \fi
  \setlength{\itemsep}{#2\baselineskip}}}
 {\end{list}}
\usepackage{calc}

\newcommand{\CSLLeftMargin}[1]{\parbox[t]{\csllabelwidth}{\strut#1\strut}}
\newcommand{\CSLRightInline}[1]{\parbox[t]{\linewidth - \csllabelwidth}{\strut#1\strut}}

\usepackage{graphicx}
\usepackage{caption}
\usepackage{bookmark}
\IfFileExists{xurl.sty}{\usepackage{xurl}}{} 
\hypersetup{
  pdftitle={Transfer Learning for Automated Feedback Generation on Small Datasets},
  pdfauthor={Oscar Morris},
  hidelinks,
  pdfcreator={LaTeX via pandoc}}

\title{Transfer Learning for Automated Feedback Generation on Small
Datasets}

\author{
 Oscar Morris \\
  \texttt{twocap06@gmail.com} \\
}
\date{March 2025}

\begin{document}
\maketitle
\begin{abstract}
Feedback is a very important part the learning process. However, it is
challenging to make this feedback both timely and accurate when relying
on human markers. This is the challenge that Automated Feedback
Generation attempts to address. In this paper, a technique to train such
a system on a very small dataset with very long sequences is presented.
Both of these attributes make this a very challenging task, however, by
using a three stage transfer learning pipeline state-of-the-art results
can be achieved with qualitatively accurate but unhuman sounding
results. The use of both Automated Essay Scoring and Automated Feedback
Generation systems in the real world is also discussed.
\end{abstract}

\section{Introduction}\label{introduction}

Automated Feedback Generation (AFG) describes the task of providing
feedback on students' work to allow them to improve. This makes the
provision of accurate feedback critical during the process of learning
and acquiring skills and knowledge {[}1{]}--{[}3{]}. Individual feedback
is often difficult and time-consuming to provide, however, providing
feedback in a timely manner is very important to a student's learning
process.

Most previous techniques perform this task by evaluating certain
features of the text, e.g.~balance of sentence types or number of
grammatical errors {[}4{]}, {[}5{]}. However, it has been shown in the
closely related domain of Automated Essay Scoring (AES) that using
neurally extracted features, either with Long-Short Term Memory (LSTM)
models {[}6{]} or using transfermer models {[}7{]} outperforms this
approach. The other benefit to using neurally extracted features is that
the models can both learn more complex features and provide more
specific feedback. When using learned features, the model is also better
able to learn the requirements of the students' task and provide
feedback on both structure and content, which is traditionally quite
challenging. Systems such as Glosser {[}4{]} use features to provide
generic reflective questions to the student. While such feedback methods
work well in discussion settings specific feedback is more effective in
most learning environments.

The rise of the pre-trained transformer model in recent years has proved
invaluable for many fields of natural language processing, including
AES, where transformers have shown a significant improvement in
performance {[}7{]}, {[}8{]}. However, they have been relatively
underused for AFG. This is primarily due to the lack of available data
that can be used to fine-tune a pre-trained transformer model.

Transfer learning allows a model which is trained in one task to re-use
knowledge in order to boost performance in a related task {[}9{]}. Using
a transfer learning pipeline can allow traditionally data-hungry models
to perform well on small datasets {[}5{]}.

In this paper, a transfer learning pipeline is presented which allows a
model to be fine-tuned using a small dataset in order to achieve
state-of-the-art performance in the task of Automated Feedback
Generation without using manually defined features or output text.

\section{Datasets}\label{datasets}

To achieve high performance on a very small dataset, three learning
stages are used, in addition to the model's pre-trained weights.

The training pipeline is designed to use large datasets first, where the
task is further removed from the desired task, followed by smaller
datasets whose task is closer to that desired.

\subsection{Fine-Tuning}\label{fine-tuning}

First, it is useful to understand the contents of the final dataset. It
is a very small dataset, containing approximately 70 samples, 20\% of
the total data was used for evaluation and therefore only approximately
56 samples can be used for training. This is a very small dataset, but
is a realistic value for these kinds of datasets.

The primary task given to students was a long-form academic writing
task, as a short review of academic literature. While they may have been
short compared to other journal articles, they are significantly longer
than most sequences dealt with by transformer models. Traditional
attention mechanisms scale (memory wise) by \(O(n^2)\) wrt. sequence
length. This poses significant challenges when processing very long
sequences (in this dataset between approx. 10,000 and 15,000 tokens).

Previous research has been conducted into efficiently processing very
long sequences, and the sparse attention mechanism was developed for use
in the longformer models {[}10{]}. The longformer model can also be used
as an encoder-decoder model in order to both accept long sequences as
input and provide long sequences as output. It is this model that is
required for this task as it can be configured to accept sequences up to
16,384 tokens in length.

\subsection{Pre-Training}\label{pre-training}

The pre-training stage consists of two sub-stages, initially the model
is trained on abstractive summarisation using the Arxiv summarisation
dataset {[}11{]} and a similar, peer-reviewing task using the PeerRead
dataset {[}12{]}.

The Arxiv summarisation dataset contains 203,037 training samples and
the PeerRead dataset contains approximately 10,000 total peer-reviews
for use in training.

In order to speed up pre-training times, a model on the huggingface
website, trained on the Arxiv summarisation dataset was used as a
starting point\footnote{https://huggingface.co/AlgorithmicResearchGroup/led\_base\_16384\_arxiv\_summarization}.

\section{Experimental Setup}\label{experimental-setup}

As previously stated, the model used is a longformer encoder-decoder
(LED) model able to efficiently process the very long sequences
contained within the datasets.

The model was trained on the Arxiv summarisation dataset for a single
epoch, other hyperparameters were not reported.

It was then trained on the PeerRead dataset for another 18 epochs using
the AdamW optimiser {[}13{]}.

Following the pre-training it was trained on the fine-tuning dataset for
50 epochs, also using the AdamW optimiser and a learning rate of
\(5 \times 10^{-5}\).

Throughout training, the model was evaluated at the end of each epoch on
the evaluation split of the relevant dataset. During the evaluation, the
ROUGE metric suite {[}14{]} is used to evaluate model performance and
are reported below.

By logging these metrics, a good picture can be built up around how the
model is performing in a quantitative way. The model is also evaluated
in a qualitative way using the fine-tuning dataset's evaluation split.

\section{Results}\label{results}

Table \ref{tbl:results} shows the evaluation data after each training
stage. When only training on the fine-tuning dataset the model was
unable to learn and perform in any meaningful way so those results are
not reported.

\begin{table}[h!]
\centering
\caption{Evaluation results on the evaluation split of each dataset. \label{tbl:results}}
\begin{tabular}{|l|llll|}
\hline
Dataset     & ROUGE-1 & ROUGE-2 & ROUGE-L & ROUGE-LSUM \\ \hline
Arxiv       & 0.37    & 0.11    & 0.20    & 0.33       \\
PeerRead    & 0.30    & 0.067   & 0.16    & 0.25       \\
Fine-tuning & 0.40    & 0.15    & 0.25    & 0.37       \\ \hline
\end{tabular}
\end{table}

As the results show the model performs well on the fine-tuning dataset
when pre-trained on first the Arxiv and then the PeerRead datasets. As
the model was unable to learn when only trained on the fine-tuning
dataset it shows very significant improvment over a simpler training
pipeline without making use of transfer learning.

Below an example automatically generated feedback is discussed,
alongside the correspending human-written feedback.

The following is the predicted feedback for a certain evaluation sample.

\begin{quote}
The abstract is well-written and gives an overview of the article.
Overall, it is a clearly laid out review that tells the story of CSP and
how this information can be inferences for companies in the area or at
the same time help understand the content. The conclusion needs to be
clearer -- more specifically, what CSP methods are and how they are used
by scientists in the Cambridge Crystallographic Data Centre (CCDC) and
their subsequent sequent tests. The wording could be improved -- some
sections have clear language, others do not. Specific comments: Page 1:
As mentioned above, the introduction seems unnecessarily vague/technical
at times. Suggestions for improvement have been provided in the
reviewer's comment form. Scope of the literature covered is broad.
Inconsistencies in the referencing style in many places and the list.
Conclusions overall were fine but less so in the way presented. In
summary, the article is well written and easy to read. But contains
several typos and grammatical mistakes -- such as ``dFT'',
``trisectational energies'' or ``etc''. As discussed, the article has a
good flow overall although some formatting inconsistencies such as use
of force fields vs force field on page 4 instead of Cornish timeliness.
Specific comments: Page 3: It would be useful to provide a concise
summary of the text. Some sentences have grammatical errors (e.g.~1,
1.1, and 1.1.1). Consistently low computational cost; Atleast one typo
per sentence (verb before last name).
\end{quote}

The following is the human-written feedback for the same evaluation
sample.

\begin{quote}
The abstract gives a good overview of this review article. Good
abstract. Introduction: The introduction briefly introduces Crystal
Structure Prediction (CSP) and CCDC. Overall good introduction.
Body/Science: The article explains the concepts involved in the review
very well step by step. Also discusses the challenges and issues
involved. Good figures used to explain body and science. Also briefly
discusses machine learning in CSP. Conclusions: The overall conclusion
is good and indicates existing gaps in knowledge and highlights
challenges and future prospects of CSP. The article has a good flow
overall and good structure. Minor typographical errors. Suggestions have
been provided in the reviewer's comment form. Scope of the literature
covered is broad. Consistent in the referencing style the list but not
in the text, use commas (with no spaces) between numbers. Suggestions
have been provided in the reviewer's comment form.
\end{quote}

The model is able to generate reasonably accurate feedback on both the
content and the structure of the review, however, some post-processing
would be useful to remove some grammatical mistakes in the generated
text. The fact that the model has learned to predict similar ideas to
those pointed out by the human marker does provide hope for a completely
neural, sequence-to-sequence approach to automatic feedback generation
on very small datasets.

The model is able to perform similarly on a variety of training samples,
covering several domains within Chemistry and in some cases is able to
provide more specific feedback than that provided by the human markers.

\section{Discussion}\label{discussion}

The use of artificial intelligence in everyday life has recently
exploded with the release of Open-AI's Chat-GPT and other openly
available large language models (LLMs). This means it is much easier
than ever before for these tools to be used by the average person. It is
tough to provide exact ethical guidance for the use of both AES and AFG
tools especially due to their use in education.

To compare AES and AFG systems, if AES systems are used by an academic
institution (e.g.~national exam boards or universities) in order to
provide final grades to a student's work which is used to partially
determine their final grade and therefore, potentially, future carreer
choices and opportunities. Since AFG is not nearly so qualitative about
a student's work and therefore is more likely to be used by the student
as a learning tool this concern is waived somewhat.

The problem of hallucinations, popularised by its noticability in LLMs
can also occur in all text models, including both AES and AFG systems.
Naturally it follows that hallucination is more of a problem in AES
systems, however, if student's learning is adversely affected by a
hallucinating AFG system, it could also have a detrimental impact on the
student's final grades. The counterpoint to the previous is that a
teacher could similarly teach incorrect information or provide incorrect
feedback when compared to their peers.

These AFG and AES systems could also be used in a semi-automatic way, in
which case they are used alongside human markers in an attempt to reduce
the workload on the human markers without sacrificing quality or the
`human touch'. A challenge with this is simply it is likely that the
human markers will over-rely on the tools they are given, this can
already be seen with openly available LLMs and it is unlikely this would
change for any useful AFG or AES system.

It has also been shown for AES systems that they can be fooled {[}15{]}.
This becomes a significant problem if these systems are used in academic
institutions as important grades may be accidentally or deliberately
inflated. It is likely that the same is true for AFG systems,
potentially creating wholly inaccurate results.

\section{Conclusion}\label{conclusion}

In this paper, a technique for the training of an Automated Feedback
System for use on small datasets is presented. The three stage training
pipeline used performs well on the small, 70 sample fine-tuning dataset
used achieving state-of-the-art results both quantatively and
qualitatively when compared with other models performing a similar task.

However, the ability to generate accurate feedback to very long text
sequences still proves challenging. Qualititavely, the model's outputs
do not read particularly well, however, it is possible that further
research or other text generaion techniques can improve this.

While it is shown that a system can perform well with extremely limited
training data (assuming other transfer learning stages are included),
these models will always perform better with larger datasets on the
desired task.

A discussion regarding the use of these techniques in the real world is
also had. Generally AFG systems are less concerning that most AES
systems due to their ability to affect a student's life. However, it
should be noted that both AES and AFG techniques prove very useful tools
in education.

\section*{References}\label{references}
\addcontentsline{toc}{section}{References}

\phantomsection\label{refs}

\end{document}